\documentclass[onecolumn]{IEEEtran}
\usepackage{amsmath,amsfonts,amssymb}
\usepackage{bm}
\usepackage{graphicx}
\usepackage{algorithm} 
\usepackage{algpseudocode} 
\usepackage{soul}
\usepackage[table,xcdraw]{xcolor}
\usepackage{multirow}
\usepackage{subcaption}
\usepackage{dsfont}

\begin{document}
\title{Evaluation and Analysis of Deep Neural Transformers and Convolutional Neural Networks on Modern Remote Sensing Datasets}

\author{J.~Alex~Hurt,
    Trevor~M.~Bajkowski,
    Grant~J.~Scott,
    and
    Curt~H.~Davis
}

\maketitle

\begin{abstract}

    In 2012, AlexNet established deep convolutional neural networks (DCNNs) as the state-of-the-art in CV,
    as these networks soon led in visual tasks for many domains, including remote sensing.
    With the publication of Visual Transformers, we are witnessing the second modern leap in computational vision, and as such, it is imperative to understand how various transformer-based neural networks perform on satellite imagery.
    While transformers have shown high levels of performance in natural language processing and CV applications, they have yet to be compared on a large scale to modern remote sensing data.
    In this paper, we explore the use of transformer-based neural networks for object detection in high-resolution electro-optical satellite imagery,
    demonstrating state-of-the-art performance on a variety of publicly available benchmark data sets.
    We compare eleven distinct bounding-box detection and localization algorithms in this study, of which seven were published since 2020, and all eleven since 2015.
    The performance of five transformer-based architectures is compared with six convolutional networks on three state-of-the-art open-source high-resolution remote sensing imagery datasets ranging in size and complexity.
    Following the training and evaluation of thirty-three deep neural models, we then discuss and analyze model performance across various feature extraction methodologies and detection algorithms.
\end{abstract}

\IEEEpeerreviewmaketitle

\section{Introduction}
\label{sect:intro}

Machine learning and computer vision (CV) have seen the incredible rise of deep neural networks (DNNs), particularly convolutional neural networks, since the original AlexNet \cite{alexnet} paper in 2012.
Combined with the processing power of GPUs and the increasing availability of robust pre-trained weights derived from massive image-sets for techniques like transfer learning, the ability to effectively train deep models has in turn led to DNNs becoming the most widely used CV technique.
In recent years, however, the convolutional feature extractors that have long been the foundation of these DNNs have been outperformed on CV challenge datasets such as the ImageNet \cite{deng2009imagenet} and COCO \cite{coco} competitions by a newer feature extraction architecture, known as transformers.

Convolutional-based deep neural networks have historically shown outstanding performance in CV applications, however, the recent publication of Visual Transformer Neural Networks, beginning with the original Vision Transformer (ViT) \cite{vit}, has enabled a leap in computational vision capabilities.
Following the publication of ViT, visual transformer architectures have been found to be capable of outperforming traditional convolutional networks for a variety of CV applications.
Given the typically ground-based perspective seen in most CV evaluation datasets, it is imperative to understand how various transformer-based neural networks perform on satellite imagery.
While transformers have shown high levels of performance in natural language processing and CV applications, they have yet to be broadly compared to CNNs on modern electro-optical (EO) high-resolution remote sensing imagery (HR-RSI).
This comparison is essential because, oftentimes, algorithms and methodologies applied in ground photo CV applications do not effectively translate to HR-RSI applications, as the unique characteristics of HR-RSI present new challenges to these emergent architectures.

Additionally, transfer learning, has been shown to provide an incredible benefit to deep neural models (see \cite{CGINets}), especially when in-domain data is limited or unavailable.
However, most readily-available pretrained weights are trained on ground photo datasets like ImageNet \cite{krizhevsky2017imagenet} and COCO \cite{coco}, which while robust, are not necessarily the most applicable dataset for transfer learning in the remote sensing (RS) domain.
By training models on popular convolutional and transformer-based models on overhead benchmark datasets, we not only enable a comparison of their performance in the RS domain but also simultaneously generate weights that can be used to improve the capabilities of future overhead CV applications via transfer learning.
These pretrained weights can be utilized to train models that perform HR-RSI tasks with a high level of performance that would not be possible without the domain-relevant pretrained weights.
To that end, all final weights derived from the experiments for this study have been made available to the scientific community. %

To effectively evaluate transformer architectures on overhead imagery and generate pretrained weights for transformer architectures in the RS domain, an incredible amount of compute is also required.
Research has shown that transformer architectures require more training time and data to effectively train compared to their convolutional siblings \cite{khan2022transformers}.
Additionally, to effectively evaluate transformers in the RS space, we must train a breadth of architectures utilizing differing feature extraction and detection techniques.
Further, to derive conclusions about the performance characteristics of any single architecture, we need to train models utilizing multiple datasets of varying sizes.

For this research, we compare eleven distinct bounding-box detection and localization algorithms, of which seven were published since 2020, and all ten since 2015.
The performance of five transformer-based architectures is compared with six convolutional networks on three state-of-the-art open-source high-resolution HR-RSI imagery datasets ranging in size and complexity.
We also investigate a diverse set of transformer architectures, including models that utilize a transformer backbone with a convolutional detection algorithm as well as end-to-end transformer detection models.
The inclusion of such a diverse set of models and methodologies, which have shown incredible performance in the ground-photo CV space, will enable meaningful takeaways and conclusions on how newer transformer architectures translate to remote sensing applications.

The rest of this paper is organized as follows.
In Sects.~\ref{sect:convnets} and \ref{sect:transformers}, we describe the deep network architectures considered for this study, including their publication dates and novel contributions.
Sect.~\ref{sect:datasets} details the publicly available remote sensing datasets that are used for experimental evaluation and their respective characteristics including their classes, relative sizes, and anticipated challenges for our networks.
Then, in Sections~\ref{sect:experimental_results} and \ref{sect:analysis}, we present experimental design, results, and analysis of the models' performance on the selected datasets.
Finally, in Sect.~\ref{sect:conclusion}, we review our conclusions and describe our future work for this research.

\section{Convolutional Networks}
\label{sect:convnets}

\begin{table*}[!t]
    \centering
    \caption{Comparison of Investigated Detection Methodologies}
    \label{table:networks}
    \def\arraystretch{1.2}
    \begin{tabular}{|l|c|c|c|c|c|c|}
        \hline
        \textbf{Detector} & \textbf{Type}    & \textbf{Backbone} & \textbf{Parameters (M)} & \textbf{$AP_{COCO}$} & \textbf{Release Year} \\
        \hline
        \hline
        ConvNeXt          & Two-Stage CNN    & ConvNeXt-S        & 67.09                   & 51.8\footnotemark[1] & 2022                  \\
        SSD               & Single-Stage CNN & VGG-16            & 36.04                   & 29.5                 & 2016                  \\
        RetinaNet         & Single-Stage CNN & ResNeXt-101       & 95.47                   & 41.6                 & 2017                  \\
        FCOS              & Single-Stage CNN & ResNeXt-101       & 89.79                   & 42.6                 & 2019                  \\
        YOLOv3            & Single-Stage CNN & DarkNet-53        & 61.95                   & 33.7                 & 2018                  \\
        YOLOX             & Single-Stage CNN & YOLOX-X           & 99.07                   & 50.9                 & 2021                  \\
        \hline
        \hline
        ViT               & Transformer      & ViT-B             & 97.62                   & N/A\footnotemark[2]  & 2020                  \\
        DETR              & Transformer      & ResNet-50         & 41.30                   & 40.1                 & 2020                  \\
        Deformable DETR   & Transformer      & ResNet-50         & 40.94                   & 46.8                 & 2020                  \\
        SWIN              & Transformer      & SWIN-T            & 45.15                   & 46.0                 & 2021                  \\
        CO-DETR           & Transformer      & SWIN-L            & 218.00                  & 64.1                 & 2023                  \\
        \hline
    \end{tabular}
\end{table*}

We begin our description of the neural architectures used in this study with the selected convolutional neural networks.
Convolutional neural networks (CNNs) have been used for CV tasks since 1989, when LeCun et al. used a CNN for handwritten digit recognition \cite{lecun1989handwritten}.
Since their inception, CNNs have grown in popularity and ability as new techniques have been created to help improve CNN performance.
These methods differ in publishing years, styles, and techniques of feature extraction and detection, and, as we will see in the experimental results section, performance characteristics on overhead imagery.
Deep CNNs have previously shown superior performance on overhead imagery, compared to hand-crafted feature extraction techniques, and are a well-suited methodology for remote sensing research in which billions of pixels are collected from remote sensing platforms each day.
This performance on overhead imagery has been demonstrated repeatedly in recent years, including in chip classification \cite{grsl_enhanced_fusion}, object detection \cite{jstars_small_object}, and semantic segmentation \cite{jstars_shapeformer}.
The key characteristics of the CNNs and Transformers compared in this study are shown in Table~\ref{table:networks}.

In the sections below, we describe the six convolutional object detection architectures we employ for this study.

\subsection{Faster R-CNN}
In 2015, Ren et. al developed the Faster R-CNN \cite{faster_rcnn} architecture for object detection.
This network was an improvement on the previous Fast R-CNN method \cite{fast_rcnn}, which required input of previously calculated region proposals to function.
The Faster R-CNN developed a method of learning the necessary region proposals from the more course set of learned feature maps using a neural network known as the Region Proposal Network (RPN).
Additionally, the RPN, Fast R-CNN, and convolutional backbone could all be trained at the same time, enabling a single training process in which only a set of images and their bounding box labels could generate a fully trained deep neural object detector.
These improvements are made possible by tiling the most course feature map with boxes called \textit{anchor boxes}, which could then be used to learn region proposals that would be used alongside Fast R-CNN to generate bounding box predictions.
The ability of Faster R-CNN to use any feature extractor, including Feature Pyramid Networks (FPN) \cite{fpn}, has allowed this architecture to remain a viable and competitive object detector.
To that end, in this study we employ a Faster R-CNN with the ConvNeXt \cite{convnext} feature extractor.
ConvNeXt is a backbone architecture that was designed to simulate transformer-like operations using standard convolutional feature extraction methods via large convolutional kernels.

\subsection{SSD}
The Single Shot Detector (SSD)\cite{ssd} was developed in 2016 as a real-time single-stage object detector.
SSD provided performance at or above that of Faster R-CNN, but with much lower latency than the popular two-stage detectors used at the time.
The unique contribution of SSD is the modification of the MultiBox method \cite{multibox} for multi-class object localization, and the real-time performance enabled by the authors' single-stage approach.
SSD functions by deriving detections directly from varying scales of feature maps throughout the convolutional layers using default boxes, a concept very similar to that of the Faster R-CNN's anchor boxes.
The selected backbone for SSD in this study is a pretrained version of the original VGG-16 \cite{vgg_net} feature extractor.

\subsection{RetinaNet}
RetinaNet \cite{retinanet} was released in 2017 by researchers at Facebook AI Research (FAIR).
This object detection architecture was designed specifically for dense object detection tasks, i.e., cars in a crowded parking lot.
Similar to SSD, RetinaNet is a single-stage object detection architecture that derives its bounding box localizations directly from the learned set of feature maps.
Unlike SSD, however, RetinaNet employs a Feature Pyramid Network (FPN) \cite{fpn} to better enable multi-scale object detection.
Additionally, the invention and use of Focal Loss is used in RetinaNet to allow for better detection in dense environments.
RetinaNet allows for a swappable convolutional feature extractor, and for this study, we choose to use an ImageNet pretrained ResNeXt-101 \cite{resnext} feature extractor.

\subsection{YOLOv3}
Perhaps the most popular object detection architecture considered here, You Only Look Once Version 3 (YOLOv3) was released in 2018 by Redmon et al. \cite{redmon2018yolov3}, and is designed mainly for real-time object detection tasks.
Unlike the other DNN architectures discussed thus far, YOLOv3 uses a custom backbone architecture similar to ResNet-50 known as DarkNet53 which utilizes residual connections between convolutional layers.
Following feature extraction, the same feature maps are then used for detection, similarly to SSD and other single-stage object detection methodologies.
YOLOv3 utilizes three parallel layers for detection at small, medium, and large object scales.

\subsection{FCOS}
The Fully Convolutional One-Stage Object Detector (FCOS) was proposed in 2019 by Tian et al \cite{fcos}.
FCOS differs from previously discussed object detection methods in that it does not utilize anchor boxes for detection; the optimization of the anchor box hyperparameters has been shown to have an impact on detection performance.
By performing detection without anchor boxes, FCOS not only removes the need to optimize those parameters but also removes the computational cost associated with calculating and learning anchor boxes.
The feature extractor used in FCOS is arbitrary and can be swapped, however, the original architecture proposed using a FPN based feature extraction with a set of detection heads for each level of the feature pyramid.
In keeping with the original publication we use a ResNeXt-101 FPN backbone \cite{resnext} for this study.

\footnotetext[1]{51.8\% COCO AP was achieved with a different detector, see \cite{convnext}}
\footnotetext[2]{ViT was not trained on COCO in the original paper}

\subsection{YOLOX}
The newest convolutional architecture considered here, YOLOX \cite{yolox} was released in 2021 by Ge et al.
YOLOX removes the anchor-based detection methodology used by other YOLO architectures, similar to the previously discussed FCOS.
Additionally, whereas previous YOLO architectures utilized coupled heads for regression and classification, YOLOX utilizes de-coupled regression and classification to enable better detection performance.
Data augmentation is also utilized heavily in training YOLOX, with methods such as MixUp and Mosaic augmentations employed during the initial exploration phase of the training process.
Finally, YOLOX employs model scaling, similar to the more recently released YOLOv5 \cite{yolov5}, enabling the creation of five different YOLOX models with different memory and compute requirements, as well as differing performance characteristics.
For the purposes of this study, we will utilize the largest and most computationally expensive version of the YOLOX architecture: YOLOX XL.
To assist this model in training, we will employ transfer learning with pretrained COCO weights for weight initialization.

\section{Transformers}
\label{sect:transformers}
We now move to the discussion of the neural transformer architectures used in this study.
Transformers have seen a rise in popularity since they were introduced to CV by the Vision Transformer (ViT) in 2020 \cite{vit}.
Since the original ViT, various other transformer models have been published, some of which utilize transformers only for feature extraction, similar to ViT, and others that instead perform CV end-to-end with transformers, such as the DETR architecture \cite{detr}. %
Unlike CNNs, transformers lack inductive spatial biases towards locality, and this distinction leads to the data-hungry nature of transformers, i.e. performing worse thatn CNNs on smaller datasets and performing better on larger datasets \cite{vit}.
Below, we describe the five transformer architectures investigated in this study for object detection, which will be compared directly with their convolutional-based siblings.
Note that for this study, when Transformer and Convolutional components are combined, such as a Transformer encoder and Convolutional localization algorithm, we consider the network overall to be a transformer network.
In other words, the use of transformer architectures in any stage of the network will designate the entire network as a transformer for the purposes of this study.

\subsection{Vision Transformer (ViT)}
The first popular transformer architecture built for CV tasks on 2-D images, the Vision Transformer (ViT) was published by researchers at Google in 2020 \cite{vit}.
The architecture was built to mirror the design of the original transformer proposed in 2017 \cite{vaswani2017attention}, but with improvements from later language-based transformer models such as BERT \cite{bert}.
Images are first converted to patch embeddings, which are 16$times$16 patches of the input imagery with positional and class embeddings to encode the positions of each patch and the appropriate class of the image being considered.
These 1-d patch embeddings are then fed to a transformer encoder identical to the original transformer (i.e., \cite{vaswani2017attention}), which consists of multi-head attention and MLP layers.
Finally, the now encoded features are fed to an MLP for final classification, as in traditional CNN classification architectures.

While vision-enabled transformers existed prior to ViT (i.e., \cite{cordonnier2019relationship}), these models had shortcomings in terms of image resolution capabilities and were unable to outperform convolutional-based architectures on common open source datasets, which held them from widespread adoption in CV. %
Conversely, ViT showed the ability of transformer-based architectures to outperform CNNs in image classification tasks, with an accuracy of 88.55\% on the ImageNet dataset, 0.15\% better than the leading convolutional network at that time.
While originally designed for classification, in this study we use the ViT architecture as a pretrained feature extractor for object detection using the RetinaNet architecture for the bounding box localization and classification task \cite{retinanet}.

\subsection{SWIN}
The SWIN Transformer is a general-use feature extraction network released by Microsoft in 2021 \cite{swin}.
This network was designed as an improvement over the previously published ViT in both computational efficiency and model performance.
To address computational efficiency, the authors employed a shifting window methodology for self-attention as opposed to the global self-attention used in ViT, reducing the required running time from quadratic to linear with respect to input image size.
Additionally, the use of hierarchical feature extraction enabled enhanced multi-scale feature extraction capabilities.
Combined, all of these changes led to a SWIN architecture that is capable of outperforming the original ViT network in both classification on ImageNet \cite{krizhevsky2017imagenet} and detection on COCO \cite{coco}.
Finally, the SWIN transformer was released with multiple scales of the network available, offering researchers a tradeoff between model performance and computational cost.
For this research, we utilize a SWIN-T feature extractor, and the bounding box localization used with the SWIN-T feature extractor is the Faster R-CNN.

\subsection{DETR}
Unlike the ViT and SWIN transformer architecture, which are general-use CV backbones, the DEtection TRansformer (DETR) \cite{detr} performs end-to-end object detection with transformers, and does not rely on alternative bounding box localization techniques.
Introduced in 2020, the DETR architecture begins with convolutional feature extraction to generate an initial set of learned features.
From these features, positional encodings are added and then passed to transformer encoder-decoder modules.
Each output embedding from the decoder is then passed to a shared feed-forward network that performs the bounding box regression and class prediction.
The authors compared DETR to Faster R-CNN, and found that DETR was able to achieve superior AP and AP50 on the popular COCO dataset, but no research has yet investigated how DETR performs on common overhead imagery.
For this research, we utilize DETR with the ResNet-50 FPN backbone.

\subsection{Deformable DETR}
While DETR outperformed Faster R-CNN for COCO detection, there still existed the opportunity to improve model performance.
To that end, Zhu et al. proposed the Deformable DETR architecture in 2021 \cite{deformable_detr}, which utilized a newly created deformable attention module.
Inspired by deformable convolutions, the deformable attention module performs attention only on a small set of key points, irrespective of the spatial size of the feature maps.
The authors then extend deformable attention modules with multi-scale deformable attention modules to enable multi-scale learning in their newly created network.
The Deformable DETR network then replaced traditional encoder-decoder transformer attention modules with these multi-scale deformable attention modules, and then compared it with the original DETR network on the COCO dataset.
Their results showed superior AP and AP50 with less training time and higher inference FPS than the original DETR on the COCO dataset.
For Deformable DETR experimentation in this paper, we will again use the ResNet-50 with FPN backbone, which will enable a fair and direct comparison in performance on overhead imagery with the original DETR network.

\subsection{CO-DETR}
The CO-DETR detector was released in 2023 at the International Conference on Computer Vision (ICCV) \cite{codetr}.
CO-DETR was developed to address a short-coming of the original DETR network, namely the lack of queries with positive samples.
This shortcoming, the authors claim, diminishes the feature learning performance of the DETR and Deformable DETR networks.
To address this lack of positive sample queries, CO-DETR utilzes a collaborative hybrid assignment training scheme, which consists of a one-to-many label assignment that improves training efficiency.
The experimental results shown with CO-DETR reflect this improved training efficiency, as CO-DETR outperforms standard DETR and Deformable DETR networks in detection on COCO using both CNN (ResNet50) and transformer (SWIN-L) feature extraction architectures.
However, CO-DETR, like many of the networks examined here, has yet to be properly evaluated in the overhead remote sensing space.

\section{Experimental Datasets}
\label{sect:datasets}

We now describe the datasets that are used for the training and testing of our deep neural models.
All of these datasets are overhead, 8-bit RGB electro-optical datasets, however, each is selected for its unique characteristics that pose a challenge to the selected models (Sect~\ref{sect:convnets} and \ref{sect:transformers}).
We choose one small dataset (25,000 labels), one medium dataset (250,000 labels), and one large-scale dataset (1 million labels), so that we may evaluate our models and their ability to generalize on different sizes and styles of overhead imagery datasets.

\begin{figure*}[!t]
    \centering
    \begin{tabular}{ccc}
        \includegraphics[width=0.3\linewidth]{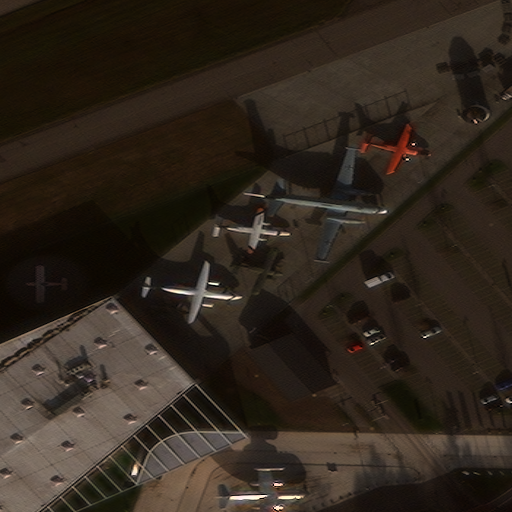} & \includegraphics[width=0.3\linewidth]{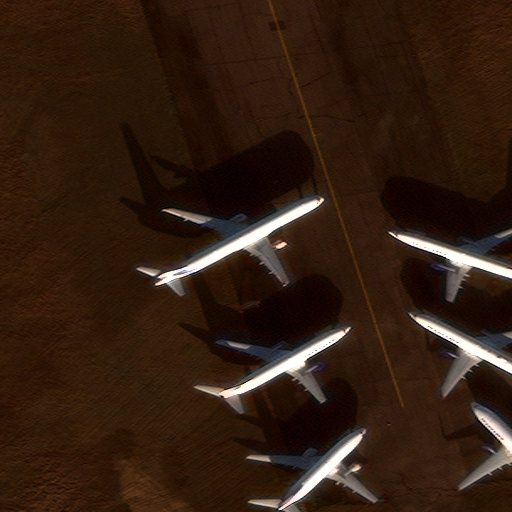} & \includegraphics[width=0.3\linewidth]{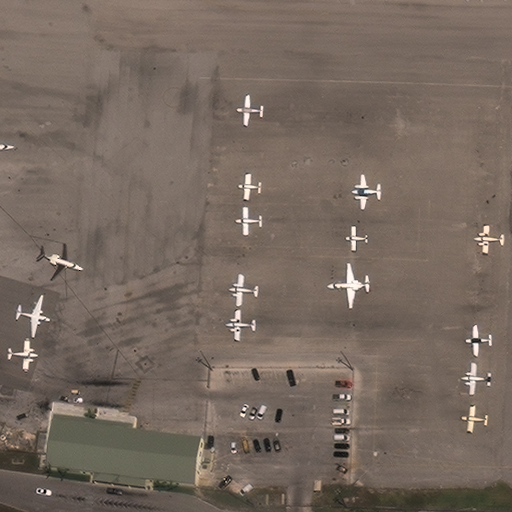} \\
                                                                                                  & \textbf{RarePlanes}                                                                       &                                                                                          \\
        \includegraphics[width=0.3\linewidth]{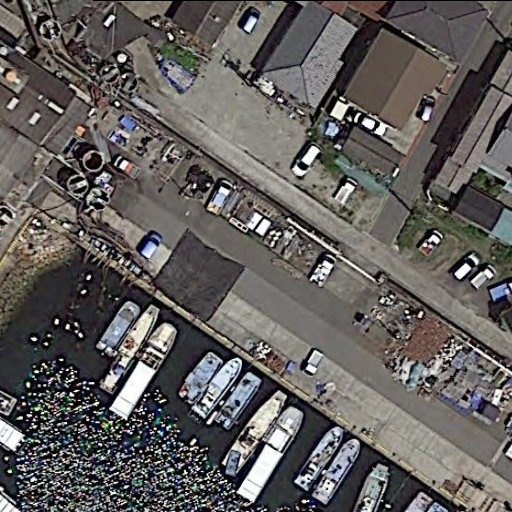}                            & \includegraphics[width=0.3\linewidth]{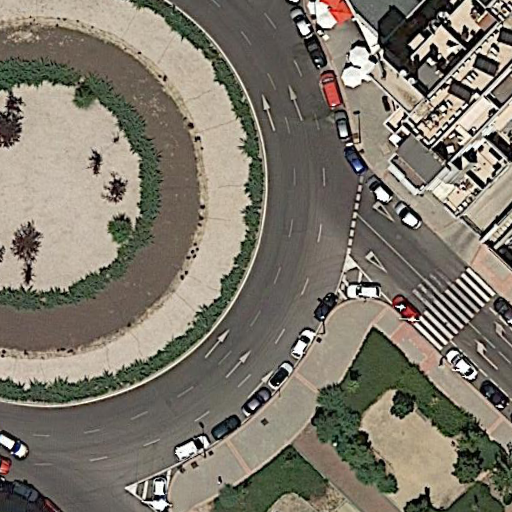}                            & \includegraphics[width=0.3\linewidth]{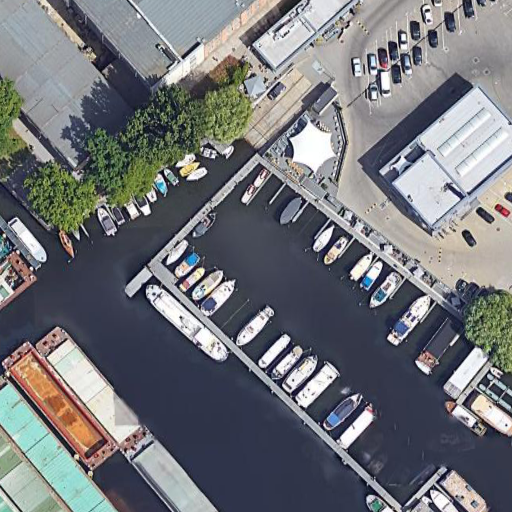}                           \\
                                                                                                  & \textbf{DOTA}                                                                             &                                                                                          \\
        \includegraphics[width=0.3\linewidth]{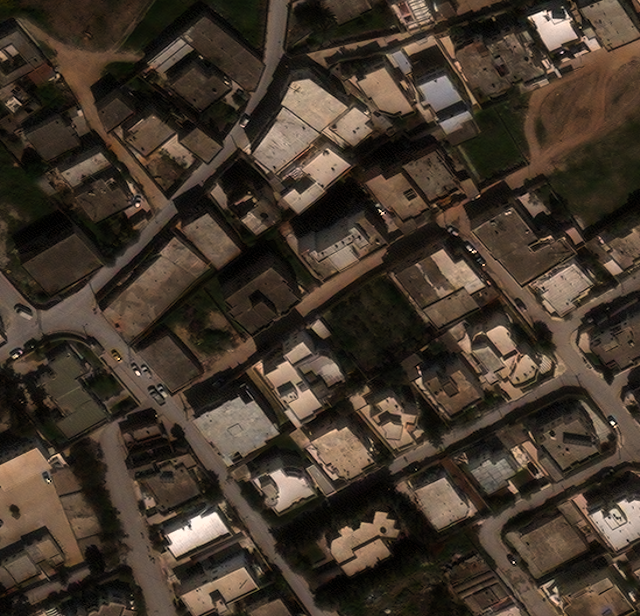}                         & \includegraphics[width=0.3\linewidth]{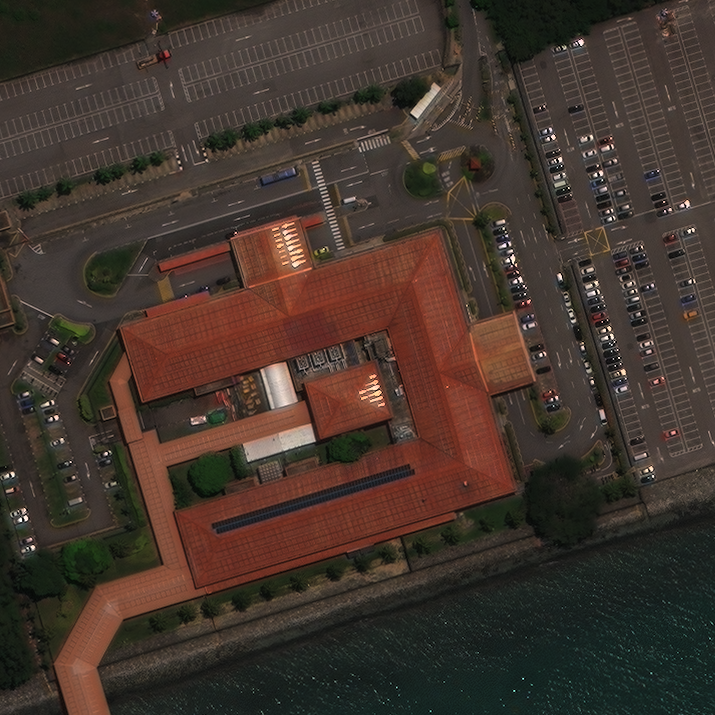}                        & \includegraphics[width=0.3\linewidth]{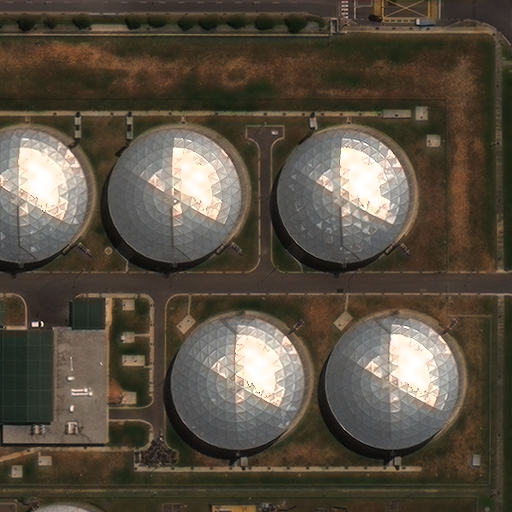}                          \\
                                                                                                  & \textbf{xView}                                                                            &                                                                                          \\
    \end{tabular}
    \caption{Sample Scenes from the RarePlanes \cite{rareplanes} (top), DOTA \cite{dota} (middle), and xView \cite{xview} (bottom) dataset dataset. Note the diversity in plane type, size, and rotational orientation in the RarePlanes dataset.}
    \label{fig:detection:rareplanes}
\end{figure*}

\subsection{RarePlanes}\label{sect:datasets:rareplans}
The smallest dataset selected for this study is the RarePlanes dataset \cite{rareplanes} that was released in 2021 by AI Reverie.
This pre-chipped pan-sharpened RGB dataset is comprised of 17,050 images belonging to seven classes of aircraft, of which three are commercial aircraft and the other four are military-related aircraft.
The original dataset was designed with real and synthetic partitions of data, but for the purposes of this study, we consider only the real data portion of the dataset.
The RarePlanes dataset is divided into sets for training and testing by its authors, and we use these partitions for our DNN training and testing.
There are 5,815 images in the train partition and 2,710 images in the test partition, giving a roughly 70\%/30\% train/test split.
The total number of samples present in the real imagery portion of the dataset is just over 25,000 objects.
This dataset will serve as the smallest dataset for our models, and due to the high class disparity seen in the RarePlanes dataset, we expect models to struggle to detect the sparse military aircraft classes reliably.
Sample imagery from the RarePlanes dataset can be seen on the top row of Fig.~\ref{fig:detection:rareplanes}.

\subsection{DOTA}\label{sect:datasets:dota}
The next dataset selected for this study is the DOTA dataset \cite{dota}.
This bounding box labeled dataset is an overhead remote sensing dataset comprised of 1869 scenes with 280,196 objects belonging to 16 classes, split into train and validation partitions.
Images from the DOTA dataset are formatted as PNGs and were collected from three separate sources with all RGB imagery collected from Google Earth, while grayscale imagery is captured from the panchromatic band of the Gaofen-2 (GF-2) and Jilin-1 (JL-1) satellites.
DOTA serves as our medium-sized dataset, with a quarter of a million labeled objects, which is much larger than the $\approx$25,000 objects present in RarePlanes.%
Additionally, unlike the highly specialized classes in RarePlanes, DOTA is a general-purpose object detection dataset, with classes ranging in relative size (pixels on target) and minor class imbalance between the 16 classes.
Given the larger disparity in visual characteristics and relative size of its classes, we expect DOTA to require the most generalizable features to detect well, which will present a unique challenge to the selected networks during training.
Sample chips from the DOTA dataset can be seen in the middle row of Fig.~\ref{fig:detection:rareplanes}.

\subsection{xView}\label{sect:datasets:xview}
Our final dataset, released by DIUx in 2017, is the xView Challenge Dataset \cite{xview}.
xView is a large-scale EO overhead remote sensing dataset that was originally released as part of an object detection challenge.
The dataset contains over 1 million objects belonging to 60 in under 200 scenes.
Scenes have been split by the xView authors into train and validation partitions, with a withheld test partition used for challenge scoring.
All imagery is pan-sharpened RGB satellite imagery.
The labels in xView are often very dense with severely overlapping objects, which presents a unique challenge to object detection models that often struggle with dense object detection applications.
Additionally, xView is a heavily imbalanced dataset with hundreds of thousands of samples in the most numerous classes but less than 1,000 samples in the most sparse class, again presenting our models with learning difficulties.
Sample chips from the xView Challenge Dataset can be seen in the bottom row of Fig.~\ref{fig:detection:rareplanes}.

\section{Experimental Results}
\label{sect:experimental_results}

We now present our experimental design and results fro the 11 Deep neural architectures examined here. We will first describe our experimental design, and then present our experimental results for all three datasets examined in this study.

\begin{table*}[!t]
    \centering
    \caption{Hyper-Parameters for Trained Models}
    \label{table:hyperparams}
    \def\arraystretch{1.2}
    \begin{tabular}{|l|c|c|c|c|c|c|}
        \hline
        \textbf{Detector} & \textbf{Backbone} & \textbf{Weights} & \textbf{Optimizer} & \textbf{Initial LR} & \textbf{Batch Size} \\
        \hline
        \hline
        ConvNeXt          & ConvNeXt-S        & ImageNet-1k      & SGD                & 0.02                & 12                  \\
        SSD               & VGG-16            & COCO             & SGD                & 0.001               & 16                  \\
        RetinaNet         & ResNeXt-101       & COCO             & SGD                & 0.001               & 2                   \\
        FCOS              & ResNeXt-101       & COCO             & SGD                & 0.02                & 4                   \\
        YOLOv3            & DarkNet-53        & COCO             & SGD                & 0.001               & 16                  \\
        YOLOX             & YOLOX-X           & COCO             & SGD                & 0.001               & 6                   \\
        \hline
        \hline
        ViT               & ViT-B             & ImageNet-1k      & AdamW              & 0.00006             & 8                   \\
        DETR              & ResNet-50         & COCO             & AdamW              & 0.0001              & 4                   \\
        Deformable DETR   & ResNet-50         & COCO             & AdamW              & 0.0002              & 4                   \\
        SWIN              & SWIN-T            & ImageNet-1k      & AdamW              & 0.0001              & 4                   \\
        CO-DETR           & SWIN-L            & COCO             & AdamW              & 0.0001              & 2                   \\
        \hline
    \end{tabular}
\end{table*}

\subsection{Experimental Setup}
Prior to training and testing with the DOTA and xView datasets, preprocessing has been applied to ease GPU acceleration.
This pre-processing includes a semi-label-aware chipping algorithm that attempts to cut all scenes from their native sizes of up to 4000 $\times$ 4000 pixels to overlapping tiles of 512 $\times$ 512 pixels.
The algorithm will attempt to make cuts in locations where bounding boxes are not severed, but rather entirely contained in a single tile if possible.
The result of this process is some duplication of labels, due to the 15\% overlap used between tiles, but it ensures that all labels remain in the final dataset used for training and allows for larger batch sizes and lowers the minimum amount of GPU memory required to train and test our models.

Network parameterization varies by architecture, with hyperparameters selected to enable the most efficient learning from the set of pretrained weights.
All networks are pretrained on either COCO or ImageNet, with most using COCO pretrained weights.
All CNNs are optimized using the Stochastic Gradient Descent algorithm, while all transformers are trained using the AdamW optimizer to match the original ViT.
The batch size used for training also varies by architecture, as all models are trained on the National Research Platform's Nautilus HyperCluster, and a dynamic batch size enabled models to efficiently train on multiple types of GPU with differing VRAM available.
Training is performed for 200 epochs for each model, with learning rate degradation utilized with a gamma of 0.1 in the final 20 and 40 epochs.
All experiments are performed utilizing the open-source MMDetection \cite{mmdetection} framework, with pretrained weights coming from this framework as well.
Table \ref{table:hyperparams} details the hyperparameters used for training the models presented in this study.

\subsection{RarePlanes Results}

\begin{table}[!t]
    \centering
    \caption{Network performance on the RarePlanes dataset. Parameters (\textbf{P}) in millions is shown with Optimal F1 Score (\textbf{F1}) alongside traditional COCO metrics.}
    \label{table:rareplanes-results}
    \def\arraystretch{1.2}
    \begin{tabular}{|l|c|c|c|c|c|c|}
        \hline
        \textbf{Model}           & \textbf{P} & \textbf{F1}       & \textbf{AP}       & \textbf{AP50}     & \textbf{AR}       & \textbf{AR50}     \\
        \hline
        \hline
        ConvNeXt                 & 67         & 57.40             & 41.06             & 54.13             & 45.76             & 56.36             \\
        SSD                      & 36         & 56.04             & 35.34             & 47.33             & 39.59             & 49.85             \\
        RetinaNet                & 95         & 67.47             & 48.06             & 60.45             & 55.79             & 67.69             \\
        FCOS                     & 90         & 68.23             & 38.32             & 47.22             & 41.70             & 48.68             \\
        YOLOv3                   & 62         & 68.59             & 42.88             & 53.88             & 45.95             & 55.31             \\
        YOLOX                    & 99         & \textit{77.14}    & \underline{54.84} & \underline{66.27} & \underline{58.22} & \underline{68.71} \\
        \hline
        \textbf{CNN Avg}         & 75         & 65.81             & 43.42             & 54.88             & 47.84             & 57.77             \\
        \hline
        \hline
        ViT                      & 98         & 52.87             & 27.02             & 44.04             & 33.13             & 49.44             \\
        DETR                     & 41         & 61.90             & 40.40             & 56.10             & 45.60             & 59.10             \\
        Deform DETR              & 41         & 67.70             & 44.42             & 57.21             & 47.97             & 58.49             \\
        SWIN                     & 45         & \textbf{81.70}    & \textbf{59.04}    & \textbf{73.71}    & \textit{61.94}    & \textit{74.47}    \\
        CO-DETR                  & 218        & \underline{70.71} & \textit{56.60}    & \textit{67.95}    & \textbf{79.74}    & \textbf{97.59}    \\
        \hline
        \textbf{Transformer Avg} & 89         & 66.98             & 45.5              & 59.8              & 53.68             & 67.82             \\

        \hline
        \multicolumn{7}{l}{Note: Top score shown in \textbf{bold}, second best score shown in \textit{italics}}                                   \\
        \multicolumn{7}{l}{and third best score is \underline{underlined}.}                                                                       \\
    \end{tabular}
\end{table}

We begin our analysis of model performance with the smallest dataset considered here, RarePlanes, which can be seen in Table~\ref{table:rareplanes-results}.
The best-performing network for the detection of RarePlanes was the SWIN Transformer.
At 81.70\% Opt F1 Score, it outperforms the next closest model, YOLOX, by over 4.5\%.
Comparing the performance of transformers among themselves, we see a significant disparity in performance moving from the highly-performant SWIN transformer and the next closest transformer model, CO-DETR.
While CO-DETR is unable to match the performance of the SWIN transformer, it outperforms its nearest transformer competitor, Deformable DETR, by over 3\%.
We also note that while CO-DETR was unable to match SWIN in Optimal F1 performance, it outperforms all other models considered here in AR50 by over 20\%, showing a significant drop in False Negatives.
Finally, ViT shows unexpectedly poor performance in detecting RarePlanes despite its previous success in ground-based Computer Vision tasks, achieving an Opt F1 score of only 52.87\%, the worst of any network examined here.

Reviewing the convolutional-based networks, we observe, similarly to the transformer models, that the best-performing model, YOLOX, far exceeds the performance of the next competing model, YOLOv3.
YOLOv3, FCOS, and RetinaNet perform similarly to each other in detecting RarePlanes, with a drop-off in performance moving to the two worst-performing convolutional models, ConvNeXt and SSD.
SSD is the oldest model trained in this study and utilizes the the much older shallower VGG-16 feature extractor, so its lower performance is expected.
However, ConvNeXt, a model built to emulate the highly-performing SWIN transformer, sees the second-worst F1 score on RarePlanes.

Comparing the overall transformer and CNN performance on RarePlanes, there is a slight performance advantage in Optimal F1 for Transformers, with a mean Optimal F1 Score of 66.98\%, compared with 65.81\% for the CNN models.
AP50 then shows a slightly larger performance gap between the transformer and CNN architectures, at 4.92\%.
However, the largest disparity in CNN and transformer performance on RarePlanes is observed in the AR50 metric, as transformers perform, on average, over 10\% better in AR50 on RarePlanes.
CO-DETR in particular exhibits excellent AR50 performance at 97.59\%, which in turn indicates an incredibly low number of false negatives detected.
Overall, RarePlanes experimentation shows superior performance for transformer models over their CNN counterparts, particularly regarding false negatives given the large disparity in AR50 scores.

\subsection{DOTA Results}
\begin{table}[!t]
    \centering
    \caption{Network performance on the DOTA dataset. Parameters (\textbf{P}) in millions is shown with Optimal F1 Score (\textbf{F1}) alongside traditional COCO metrics.}
    \label{table:dota-results}
    \def\arraystretch{1.2}
    \begin{tabular}{|l|c|c|c|c|c|c|}
        \hline
        \textbf{Model}           & \textbf{P} & \textbf{F1}       & \textbf{AP}       & \textbf{AP50}     & \textbf{AR}       & \textbf{AR50}     \\
        \hline
        \hline
        ConvNeXt                 & 67         & 63.18             & 30.52             & 52.71             & 37.29             & 59.00             \\
        SSD                      & 36         & 62.17             & 24.59             & 45.39             & 31.59             & 51.62             \\
        RetinaNet                & 95         & 66.32             & 31.05             & 54.10             & 39.51             & 62.10             \\
        FCOS                     & 90         & 40.34             & 8.72              & 14.78             & 9.71              & 14.82             \\
        YOLOv3                   & 62         & 64.02             & 22.22             & 45.04             & 30.26             & 53.34             \\
        YOLOX                    & 99         & \textit{72.84}    & \textit{35.27}    & \textit{60.67}    & \textit{44.18}    & \textit{66.67}    \\
        \hline
        \textbf{CNN Avg}         & 75         & 61.48             & 25.40             & 45.45             & 32.09             & 51.26             \\
        \hline
        \hline
        ViT                      & 98         & 59.36             & 21.01             & 43.07             & 30.10             & 53.58             \\
        DETR                     & 41         & 55.40             & 11.60             & 30.70             & 21.30             & 46.10             \\
        Deform DETR              & 41         & 66.30             & 26.76             & 49.79             & 34.75             & 57.30             \\
        SWIN                     & 45         & \underline{68.87} & \underline{33.46} & \underline{58.09} & \underline{41.49} & \underline{64.55} \\
        CO-DETR                  & 218        & \textbf{73.53}    & \textbf{38.60}    & \textbf{67.25}    & \textbf{49.59}    & \textbf{79.37}    \\
        \hline
        \textbf{Transformer Avg} & 89         & 66.98             & 45.5              & 59.8              & 53.68             & 67.82             \\

        \hline
        \multicolumn{7}{l}{Note: Top score shown in \textbf{bold}, second best score shown in \textit{italics}}                                   \\
        \multicolumn{7}{l}{and third best score is \underline{underlined}.}                                                                       \\
    \end{tabular}
\end{table}

We now review the performance of our selected networks on the DOTA dataset.
Unlike RarePlanes, DOTA contains a diverse array of 16 classes, and over 250,000 total objects to detect.
The three best-performing networks in Optimal F1 Score, as shown in Table~\ref{table:dota-results}, are CO-DETR (73.53), YOLOX (72.84\%), and SWIN (68.87\%).
CO-DETR performs best not only on Optimal F1 Score on DOTA but also outperforms all other networks in all reported metrics here, showing that the CO-DETR architecture is well suited to the detection of this satellite imagery dataset.

Comparing CNN models, we see a significant drop in performance moving from the best-performing CNN, YOLOX, to the next best model, RetinaNet.
However, four of the six CNNs perform remarkably similarly on DOTA, as RetinaNet, YOLOv3, ConvNeXt, and SSD all achieve an Opt F1 score between 62\% and 67\%.
The final remaining outlier, then, is the FCOS model, which heavily struggles to detect the DOTA dataset, achieving a mere 40.34\% Opt F1 Score, the lowest of any model considered in this study.

Next, we review the relative performance of the transformer models.
Aside from CO-DETR's outstanding detection performance, we note that SWIN again performs well as the second-best transformer, but without as wide of a margin over the third-best transformer, Deformable DETR.
At 68.87\% Opt F1, SWIN sits just above the 66.3\% of Deformable DETR, which similarly to RarePlanes, outperforms its sibling DETR network by a large margin of over 10\%.
However, unlike the performance of ViT in RarePlanes, ViT is not the worst performing model on DOTA, as it outperforms DETR in DOTA detection and is only 7\% worse than the Deformable DETR model at 59.36\%.

Comparing transformer models against CNN models on DOTA again shows competitive performance, but with more recently developed transformer models having an advantage over the CNN models.
Mean Opt F1 is more than 3\% better for transformers, with that gap widening to over 4\% on AP50 and over 9.5\% on AR50, again propelled primarily by the AR50 achieved by the leading CO-DETR model.

\begin{figure*}[!t]
    \centering
    \includegraphics[width=.9\linewidth]{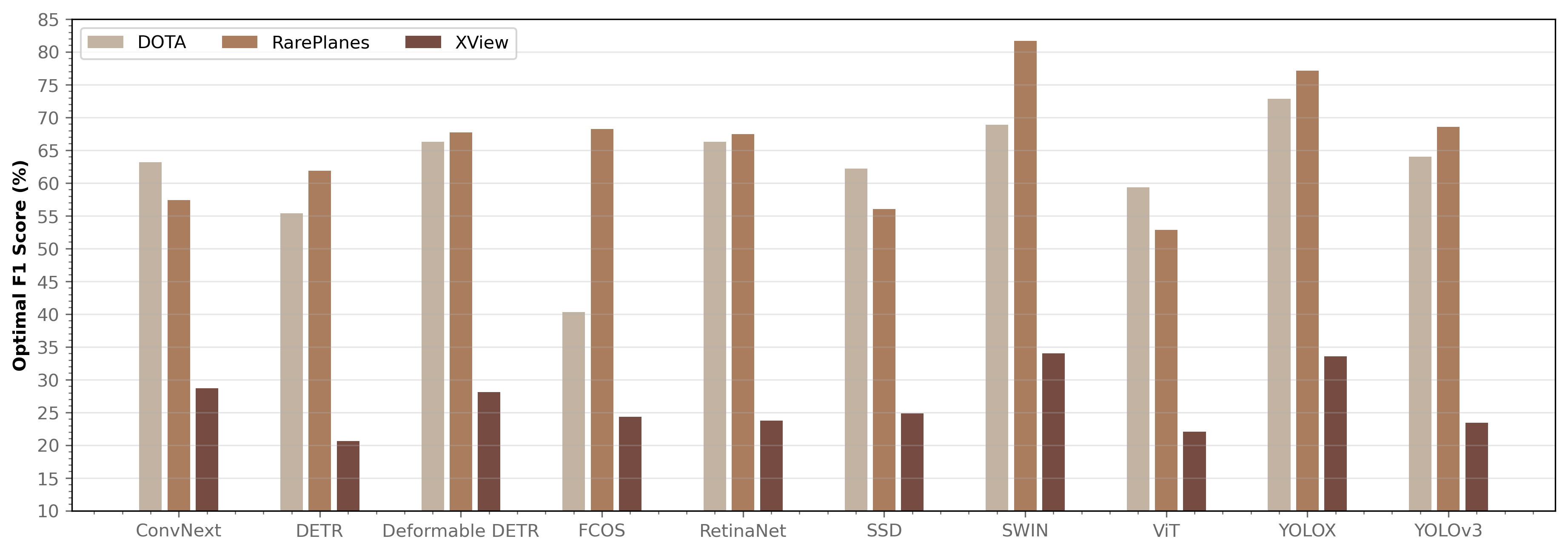}
    \caption{Comparison of Model Performance Across Datasets}
    \label{fig:dataset-compare}
\end{figure*}

\subsection{xView Results}
\begin{table}[!t]
    \centering
    \caption{Network performance on the XView dataset. Parameters (\textbf{P}) in millions is shown with Optimal F1 Score (\textbf{F1}) alongside traditional COCO metrics.}
    \label{table:xview-results}
    \def\arraystretch{1.2}
    \begin{tabular}{|l|c|c|c|c|c|c|}
        \hline
        \textbf{Model}           & \textbf{P} & \textbf{F1}       & \textbf{AP}      & \textbf{AP50}     & \textbf{AR}       & \textbf{AR50}     \\
        \hline
        \hline
        ConvNeXt                 & 67         & 28.67             & 6.47             & 13.45             & 10.70             & 19.72             \\
        SSD                      & 36         & 24.86             & 4.56             & 9.71              & 6.73              & 12.61             \\
        RetinaNet                & 95         & 23.74             & 4.97             & 10.02             & 8.00              & 14.28             \\
        FCOS                     & 90         & 24.35             & 4.07             & 7.81              & 5.71              & 9.69              \\
        YOLOv3                   & 62         & 23.44             & 3.42             & 8.11              & 5.36              & 10.78             \\
        YOLOX                    & 99         & \textit{33.57}    & \underline{8.70} & \underline{16.09} & \underline{12.67} & \underline{21.10} \\
        \hline
        \textbf{CNN Avg}         & 75         & 26.44             & 5.37             & 10.87             & 8.20              & 14.70             \\
        \hline
        \hline
        ViT                      & 98         & 22.08             & 3.81             & 8.83              & 6.29              & 12.93             \\
        DETR                     & 41         & 20.63             & 2.05             & 5.78              & 4.99              & 12.11             \\
        Deform DETR              & 41         & 28.13             & 6.59             & 13.46             & 9.79              & 17.65             \\
        SWIN                     & 45         & \textbf{34.04}    & \textit{9.01}    & \textit{17.58}    & \textit{13.30}    & \textit{23.30}    \\
        CO-DETR                  & 218        & \underline{29.19} & \textbf{9.38}    & \textbf{19.69}    & \textbf{24.19}    & \textbf{45.31}    \\
        \hline
        \textbf{Transformer Avg} & 89         & 66.98             & 45.5             & 59.8              & 53.68             & 67.82             \\

        \hline
        \multicolumn{7}{l}{Note: Top score shown in \textbf{bold}, second best score shown in \textit{italics}}                                  \\
        \multicolumn{7}{l}{and third best score is \underline{underlined}.}                                                                      \\
    \end{tabular}
\end{table}

We move now to the analysis of model performance on the xView dataset.
Recall that xView is the largest and most challenging dataset considered here, with dense object detection and imbalanced class count challenges presented to models.
Reviewing the results, shown in Table~\ref{table:xview-results}, we first note that as with RarePlanes, SWIN is the best-performing network and YOLOX is the second-best performing network.
Further, we see that all of our models have struggled to detect xView at a high level, with the best overall performance netting only 34.04\% Opt F1 Score and 17.59\% AP50, highlighting the relative difficulty in detecting the xView dataset at a high level.

Comparing the relative performance of transformer models, we see a nearly 5\% drop in Opt F1 performance from SWIN to the CO-DETR model.
CO-DETR again shows incredibly high AR and AR50 scores, achieving an AR50 score of more than 20\% better than the next best model.
Deformable DETR again shows a significant leap in performance compared to the original DETR model, with ViT again outperforming the DETR model.
Reviewing the CNN models' relative performance, we see the YOLOX is nearly 5\% better than the next CNN model ConvNeXt, which in turn is then nearly 4\% better than the next best model.
The remaining CNN models, SSD, RetinaNet, FCOS, and YOLOv3 all perform similarly with scores between 23.44\% and 24.86\%.

\section{Analysis and Discussion}
\label{sect:analysis}
We now present our larger analysis and discussion of these results. We will first examine the models' performance across datasets. We will then compare Transformer and CNN-based architectures on our datasets before discussing a case study using the only detection algorithm that was trained with both a CNN and Transformer backbone. Finally, we will analyze our models' performance in the context of their model complexity and requisite training times.

\subsection{Comparing Model Performance Across Datasets}

We begin our discussion by comparing the performance of our models across datasets.
Due to the relative difficulty of the datasets considered for this study, model performance was not consistent.
As can be seen in Figure~\ref{fig:dataset-compare}, eight of the eleven models performed best on RarePlanes, with a slightly lower score on DOTA, followed by a significant drop in performance on xView.
Three models, however, found higher scores on DOTA than on RarePlanes, including ConvNext, SSD, and ViT.

For all models, we discovered that the average difference in performance between DOTA and RarePlanes was 7.7\%, and the average difference between for each model between the best (DOTA or RarePlanes) and worst (XView) dataset was 41.39\%.
These performance gaps reinforce prior results in the DNN space that show that oftentimes, models are limited in their performance by the dataset upon which they are trained.
XView is an incredibly challenging dataset, and so it does follow that our models would struggle heavily with detecting it at a high level.
RarePlanes, on the other hand, is a more consistent, less heterogeneous, and simpler dataset to detect, and thus it follows that most models would see their best performance on this dataset.

\subsection{Comparing Transformer and CNN Performance}
Our final performance analysis examines the performance of the investigated transformer models (ViT, DETR, Deformable DETR, CO-DETR, and SWIN) against the investigated CNN models (ConvNeXt, SSD, RetinaNet, FCOS, YOLOv3, and YOLOX).
To aid this comparison, we begin by ranking the performance of our eleven models by their average Opt F1 score on each dataset and then average the ranks across models and datasets.

The results, seen in Table~\ref{table:rank-compare}, show three models as the clear and consistent best-performers: SWIN, YOLOX, and CO-DETR.
These three models, which averaged a ranking of 1.67, 2.00, and 2.33 respectively, on the three datasets, were the highest-performing models, and by significant margins in several cases.
Given that deep networks may thrive or struggle in different detection contexts based on their design philosophy, we also observe that for the remaining eight models, four of them are rather inconsistent, and see a large disparity in rank from dataset to dataset.
These five CNN models, ConvNeXt, SSD, RetinaNet, FCOS, and YOLOv3, range in inconsistency, with ConvNeXt ranking the 10/11 on RarePlanes but the 4/11 on xView.
Similarly, RetinaNet is ranked 8/11 on xView but 4/11 for DOTA, and FCOS is 5/11 for RarePlanes, but the worst model on DOTA.

With three leading performers and five inconsistent networks, we are then left with the three remaining transformer models: ViT, DETR, and Deformable DETR.
Two of these models, ViT and DETR, were consistently in the bottom 3 models for each dataset, with average Opt F1 ranks of 10 and 9.67, respectively.
The final model, Deformable DETR is a consistently average performer, with an average rank of 5.67 and scoring 5th or 7th out of 11 models on all datasets.

\begin{table}[!t]
    \centering
    \caption{Ranks of Models' Optimal F1 Score on Investigated Datasets}
    \label{table:rank-compare}
    \def\arraystretch{1.2}
    \begin{tabular}{|l|c|c|c||c|c|}
        \hline
        \textbf{Model}               & \textbf{RP} & \textbf{DOTA} & \textbf{xView} & \textbf{Avg}  & \textbf{Std} \\
        \hline
        \hline
        ConvNeXt                     & 10          & 7             & 4              & 7             & 3            \\
        SSD                          & 9           & 8             & 6              & 7.67          & 1.53         \\
        RetinaNet                    & 6           & 4             & 8              & 6             & 2            \\
        FCOS                         & 5           & 11            & 7              & 7.67          & 3.06         \\
        YOLOv3                       & 4           & 6             & 9              & 6.33          & 2.52         \\
        YOLOX                        & 2           & 2             & 2              & 2             & \textbf{0}   \\
        \hline
        \textbf{CNN Average}         & 6           & 6.33          & 6              & 6.11          & 2.02         \\
        \hline
        \hline
        ViT                          & 11          & 9             & 10             & 10            & 1            \\
        DETR                         & 8           & 10            & 11             & 9.67          & 1.53         \\
        Deformable DETR              & 7           & 5             & 5              & 5.67          & 1.15         \\
        SWIN                         & \textbf{1}  & 3             & \textbf{1}     & \textbf{1.67} & 1.15         \\
        CO-DETR                      & 3           & \textbf{1}    & 3              & 2.33          & 1.15         \\
        \hline
        \textbf{Transformer Average} & 6           & 5.6           & 6              & 5.87          & 1.20         \\
        \hline
    \end{tabular}
\end{table}

Overall, transformer performance was more consistent across the overhead remote-sensing datasets examined here with less average rank variation compared to CNNs.
Additionally, the average rank of the transformer models was slightly better (lower) than that of the CNNs for the DOTA dataset, but was equivalent for both RarePlanes and xView.
However, while the average ranks of each group may be equivalent for two datasets, the mean Opt F1 score for transformers was higher than that of CNNs for all investigated datasets here.

\subsection{Model Complexity and Training Time Considerations}
For our final analysis of our selected models, we examine the tradeoffs between model training times and model performance.
Due to the heterogeneous hardware infrastructure found on the NRP, the models presented here were trained with a large variety of hardware the GPU types, preventing any direct comparison in training times.
However, we have chosen to perform a smaller-scale experiment to measure the relative computational complexity of the networks.
For this experiment, we will train an identical version of each network on RarePlanes for 500 iterations with a batch size of 8 using only a single NVIDIA RTX 3090, utilize this timing data to calculate the number of images (frames) processed per unit of time (seconds), and then use this value to extrapolate the training times for all of our models.
This will enable direct comparisons in training times and enable discussions of the relative performance versus training time tradeoffs that researchers face in model selection for remote sensing applications.

\begin{table}[!t]
    \centering
    \caption{Comparison of each network's 500 iteration train time (in seconds), calculated images (frames) processed per second (FPS), and F1 Score on each dataset}
    \label{table:timing-compare}
    \def\arraystretch{1.2}
    \begin{tabular}{|l|c|c||c|c|c|}
        \hline
        \textbf{Model}  & \textbf{Time}  & \textbf{FPS}  & \textbf{RP}   & \textbf{DOTA}   & \textbf{xView} \\
        \hline
        ConvNeXt        & 316.04         & 12.66         & 57.4          & 63.18           & 28.67          \\
        SSD             & 106.24         & 37.65         & 56.04         & 62.17           & 24.86          \\
        RetinaNet       & 291.12         & 13.74         & 67.47         & 66.32           & 23.74          \\
        FCOS            & 178.89         & 22.36         & 68.23         & 40.34           & 24.35          \\
        YOLOv3          & \textbf{87.92} & \textbf{45.5} & 68.59         & 64.02           & 23.44          \\
        YOLOX           & 271.32         & 14.74         & 77.14         & 72.84           & 33.57          \\
        \hline
        \hline
        ViT             & 248.13         & 16.12         & 52.87         & 59.36           & 22.08          \\
        DETR            & 165.4          & 24.18         & 61.9          & 55.4            & 20.63          \\
        Deformable DETR & 408.92         & 9.78          & 67.7          & 66.3            & 28.13          \\
        SWIN            & 254.03         & 15.75         & \textbf{81.7} & 68.87           & \textbf{34.04} \\
        CO-DETR         & 796.58         & 5.02          & 70.71         & \textbf{73.53 } & 29.19          \\
        \hline
    \end{tabular}
\end{table}

Reviewing the timing results, shown alongside per dataset Mean Optimal F1 scores in Table~\ref{table:timing-compare}, we observe that there are large discrepancies between various models in terms of capable FPS throughput.
Seven of the eleven networks investigated here, including three CNNs and four transformers, achieve an FPS (on an NVIDIA RTX 3090) under 20 FPS, indicating the high computational costs associated with training modern deep networks.
Only two models, SSD and YOLOv3, manage to achieve upwards of 30 FPS, which was an explicit choice in these networks' original designs.
The three best overall models across the three datasets investigated in this research, YOLOX, SWIN, and CO-DETR, all show exceptionally high training times in the 500-iteration experiment, requiring as much as 13 minutes to process only 4,000 512 $\times$ 512 RGB images,
SWIN and YOLOX manage to net nearly 15 FPS (14.74 for YOLOX and 15.75 for SWIN), however, the best performer on DOTA, CO-DETR, manages to achieve a measly 5.02 FPS in our training benchmark.
These differences can likely be attributed to the sizes of the feature extractors more than the detection methodology itself, as CO-DETR utilizes a SWIN-L feature extractor.

\begin{figure}[!t]
    \includegraphics[width=\linewidth]{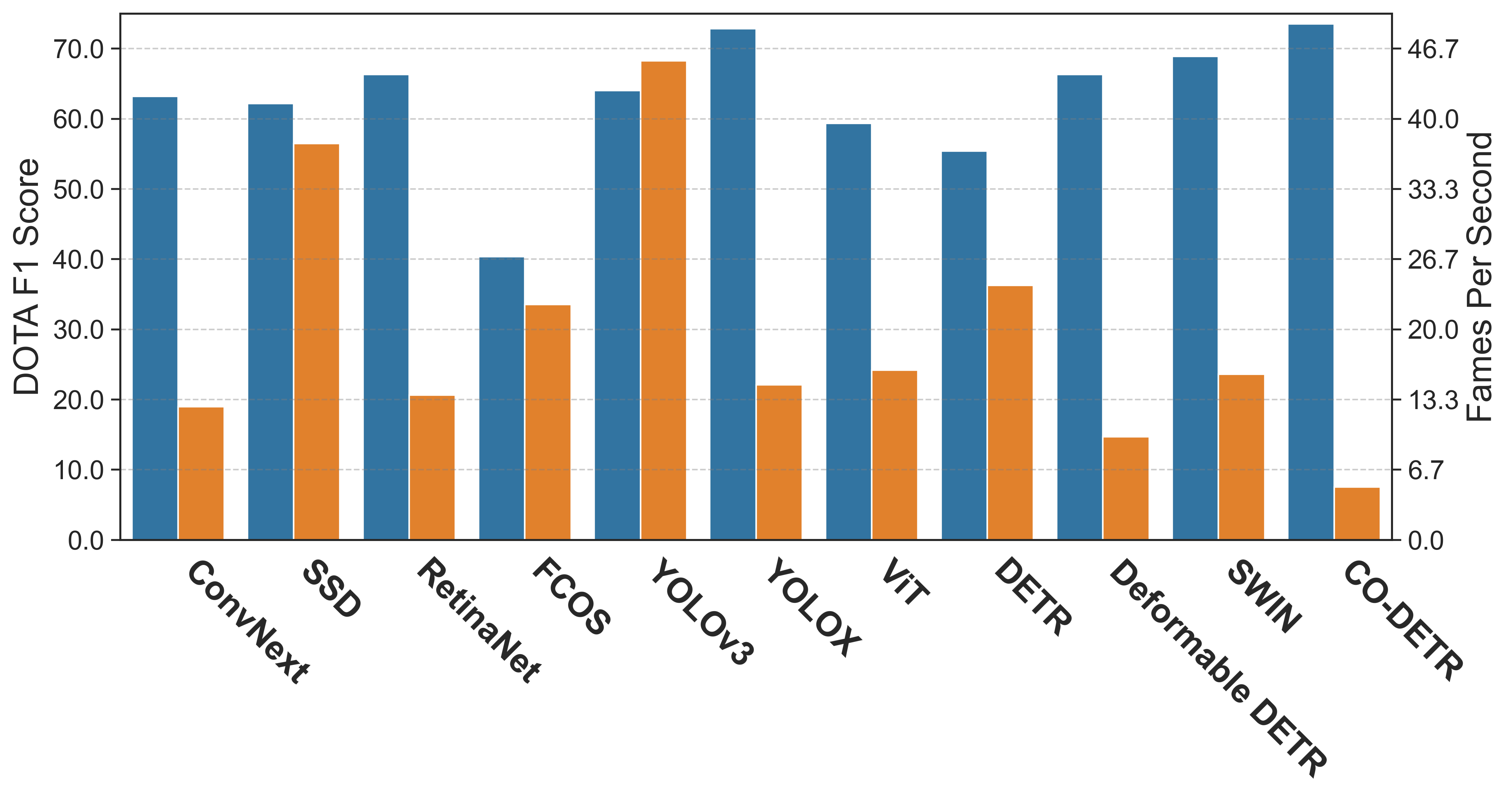}
    \caption{DOTA F1 Score (blue) versus FPS in training (orange)}
    \label{fig:timing}
\end{figure}

While analyzing overall training time in a fixed hardware environment is important for understanding a model's performance characteristics, it is far more crucial to understand the requisite training timers considering the model's performance once trained.
The importance of this analysis comes from the heterogeneous requirements found in remote sensing applications.
Some applications of DNNs in the HR-RSI space may prioritize model performance over training throughput while others may have limited resources and require the selection of a less computationally expensive model at the cost of performance.
In an effort to better understand these tradeoffs on a per-network basis on remote sensing data, we co-plot the FPS training throughput of each of our eleven models alongside the F1 Score on the DOTA dataset in Fig.~\ref{fig:timing}.
We select DOTA for this analysis, as we believe it to be the most general-use dataset examined here, and thus our conclusions should better generalize across various remote sensing applications.

Our first takeaway from Fig.~\ref{fig:timing} is that the most performant dataset, CO-DETR, is also the most computationally expensive, which is expected given the choice of feature extractor (SWIN-L).
Meanwhile, the second best performer by F1 Score, YOLOX, achieves 2.94x the FPS performance while scoring only 0.69\% lower in Mean Opt F1 Score.
While there exist applications in which the additional nearly 0.7\% would be preferred, there are many applications in which YOLOX may be a more preferred network given the tradeoff between F1 score and training times.
Extending this discussion further, YOLOv3, the sixth best model on DOTA, achieves an incredibly high 45 FPS in our training benchmark, and while it is outperformed by the more modern Deformable DETR (fifth-ranked on DOTA) by nearly 5\% (4.85\%), this 5\% tradeoff in F1 Score would net a training time improvement of over 450\%.
There are again certainly applications in which a 68.87\% F1 Score would warrant the 450\% increase in training time, however, many applications may find the choice of a YOLOv3 model at 64\% F1 score acceptable in exchange for a 45 FPS network (compared to under 10 FPS).
Overall, our examination of training times against model performance shows that modern DNN practitioners have the benefit of selecting their optimal operating point between training time and model performance using either CNN or Transformer-based networks.

\subsection{Case Study: RetinaNet}

We now move to a case study using RetinaNet.
RetinaNet is the only detection algorithm examined here that was trained with both a convolutional and transformer feature extractor.
Additionally, these two models have a remarkably similar number of learnable parameters, with ViT utilizing 97M and ResNeXt-101 utilizing 95M parameters (Table \ref{table:networks}).
As such, comparing these models will allow to us directly compare the feature extraction performance of CNNs and Transformers.

In comparing RetinaNet with ResNeXt-101 (CNN) against the RetinaNet with ViT (Transformer), we first note the model performance on the DOTA dataset, seen in Table~\ref{table:dota-results}.
For all measured metrics, the CNN-based RetinaNet with ResNeXt-101 outperforms the Transformer-based RetinaNet with ViT, with some metrics showing a very significant gap.
The gap in Optimal F1 is 6.96\%, and the gap in the popular AP50 metric is 11.03\%, showing that for the DOTA dataset, the CNN-based ResNeXt-101 feature extractor provided suprior features for object localization and classification than the Transformer-based ViT-B.

These results hold for the smaller RarePlanes dataset, but for the larger XView dataset, we see the gap between the two RetinaNet models narrow.
As seen in Table~\ref{table:xview-results}, the ViT-B model performs much closer to the RetinaNet, suggesting that the larger dataset is better suited for learning with ViT.
These results reinforce previous publications that show that Transformers are more data hungry, and require more images to effectively train for the same number of learnable parameters.

It should also be noted that while CNN-based RetinaNet outperformed the Transformer-based RetinaNet in model performance, the ViT-B model is less computationally expensive.
Based on our testing shown in Table~\ref{table:timing-compare}, ViT is 14.8\% faster than ResNeXt-101 in training with RetinaNet.

\section{Conclusion}
\label{sect:conclusion}

Transformer models have shown great promise in the ground-based CV space, but to this point, have not been comprehensively evaluated in overhead object detection with EO imagery.
We have shown numerous experimental results of five state-of-the-art transformers, including the original ViT and ubiquitous SWIN transformer, and compared their performance against six convolutional-based deep neural network object detectors on three overhead, EO high-resolution object detection datasets.
These three datasets, RarePlanes, DOTA, and xView, range in intra-class diversity, class sizes, number of samples, and visual characteristics, and each presents a unique challenge to our models.

Our results show that both transformer models and CNNs are able to perform object detection on high-resolution EO imagery at a competitive and highly performing rate, with the best transformer models providing superior performance on all datasets in both Opt F1 and the standard COCO metrics.
However, this superior performance came at the cost of longer training times and lower real-time FPS performance.
Our analysis also showed that the transformers perform more consistently across EO datasets, while several of our CNN models displayed highly inconsistent detection performance moving between overhead imagery datasets.

\section*{Acknowledgement}

Computational resources for this research have been supported by the NSF National Research Platform, as part of GP-ENGINE (award OAC \#2322218).

\bibliographystyle{IEEEtran}
\bibliography{refs}

\begin{thebibliography}{10}
\providecommand{\url}[1]{#1}
\csname url@samestyle\endcsname
\providecommand{\newblock}{\relax}
\providecommand{\bibinfo}[2]{#2}
\providecommand{\BIBentrySTDinterwordspacing}{\spaceskip=0pt\relax}
\providecommand{\BIBentryALTinterwordstretchfactor}{4}
\providecommand{\BIBentryALTinterwordspacing}{\spaceskip=\fontdimen2\font plus
\BIBentryALTinterwordstretchfactor\fontdimen3\font minus \fontdimen4\font\relax}
\providecommand{\BIBforeignlanguage}[2]{{%
\expandafter\ifx\csname l@#1\endcsname\relax
\typeout{** WARNING: IEEEtran.bst: No hyphenation pattern has been}%
\typeout{** loaded for the language `#1'. Using the pattern for}%
\typeout{** the default language instead.}%
\else
\language=\csname l@#1\endcsname
\fi
#2}}
\providecommand{\BIBdecl}{\relax}
\BIBdecl

\bibitem{alexnet}
A.~Krizhevsky, I.~Sutskever, and G.~E. Hinton, ``Imagenet classification with deep convolutional neural networks,'' \emph{Advances in neural information processing systems}, vol.~25, pp. 1097--1105, 2012.

\bibitem{deng2009imagenet}
J.~Deng, W.~Dong, R.~Socher, L.-J. Li, K.~Li, and L.~Fei-Fei, ``Imagenet: A large-scale hierarchical image database,'' in \emph{2009 IEEE conference on computer vision and pattern recognition}.\hskip 1em plus 0.5em minus 0.4em\relax Ieee, 2009, pp. 248--255.

\bibitem{coco}
T.-Y. Lin, M.~Maire, S.~Belongie, J.~Hays, P.~Perona, D.~Ramanan, P.~Doll{\'a}r, and C.~L. Zitnick, ``Microsoft coco: Common objects in context,'' in \emph{European conference on computer vision}.\hskip 1em plus 0.5em minus 0.4em\relax Springer, 2014, pp. 740--755.

\bibitem{vit}
A.~Dosovitskiy, L.~Beyer, A.~Kolesnikov, D.~Weissenborn, X.~Zhai, T.~Unterthiner, M.~Dehghani, M.~Minderer, G.~Heigold, S.~Gelly \emph{et~al.}, ``An image is worth 16x16 words: Transformers for image recognition at scale,'' \emph{arXiv preprint arXiv:2010.11929}, 2020.

\bibitem{CGINets}
G.~J. Scott, M.~R. England, W.~A. Starms, R.~A. Marcum, and C.~H. Davis, ``Training deep convolutional neural networks for land-cover classification of high-resolution imagery,'' \emph{IEEE GRSL}, vol.~14, no.~4, pp. 549--553, 2017.

\bibitem{krizhevsky2017imagenet}
A.~Krizhevsky, I.~Sutskever, and G.~E. Hinton, ``Imagenet classification with deep convolutional neural networks,'' \emph{Communications of the ACM}, vol.~60, no.~6, pp. 84--90, 2017.

\bibitem{khan2022transformers}
S.~Khan, M.~Naseer, M.~Hayat, S.~W. Zamir, F.~S. Khan, and M.~Shah, ``Transformers in vision: A survey,'' \emph{ACM computing surveys (CSUR)}, vol.~54, no. 10s, pp. 1--41, 2022.

\bibitem{lecun1989handwritten}
Y.~LeCun, B.~Boser, J.~Denker, D.~Henderson, R.~Howard, W.~Hubbard, and L.~Jackel, ``Handwritten digit recognition with a back-propagation network,'' \emph{Advances in neural information processing systems}, vol.~2, 1989.

\bibitem{grsl_enhanced_fusion}
G.~J. Scott, K.~C. Hagan, R.~A. Marcum, J.~A. Hurt, D.~T. Anderson, and C.~H. Davis, ``Enhanced fusion of deep neural networks for classification of benchmark high-resolution image data sets,'' \emph{IEEE GRSL}, vol.~15, no.~9, pp. 1451--1455, 2018.

\bibitem{jstars_small_object}
H.~Yi, B.~Liu, B.~Zhao, and E.~Liu, ``Small object detection algorithm based on improved yolov8 for remote sensing,'' \emph{IEEE Journal of Selected Topics in Applied Earth Observations and Remote Sensing}, 2023.

\bibitem{jstars_shapeformer}
P.~Lv, L.~Ma, Q.~Li, and F.~Du, ``Shapeformer: a shape-enhanced vision transformer model for optical remote sensing image landslide detection,'' \emph{IEEE Journal of Selected Topics in Applied Earth Observations and Remote Sensing}, vol.~16, pp. 2681--2689, 2023.

\bibitem{faster_rcnn}
S.~Ren, K.~He, R.~Girshick, and J.~Sun, ``Faster r-cnn: towards real-time object detection with region proposal networks,'' \emph{IEEE transactions on pattern analysis and machine intelligence}, vol.~39, no.~6, pp. 1137--1149, 2016.

\bibitem{fast_rcnn}
R.~Girshick, ``Fast r-cnn object detection with caffe,'' \emph{Microsoft Research}, 2015.

\bibitem{fpn}
T.-Y. Lin, P.~Doll{\'a}r, R.~Girshick, K.~He, B.~Hariharan, and S.~Belongie, ``Feature pyramid networks for object detection,'' in \emph{Proceedings of the IEEE conference on computer vision and pattern recognition}, 2017, pp. 2117--2125.

\bibitem{convnext}
Z.~Liu, H.~Mao, C.-Y. Wu, C.~Feichtenhofer, T.~Darrell, and S.~Xie, ``A convnet for the 2020s,'' in \emph{Proceedings of the IEEE/CVF Conference on Computer Vision and Pattern Recognition}, 2022, pp. 11\,976--11\,986.

\bibitem{ssd}
W.~L. et~al., ``{SSD}: Single shot multibox detector,'' in \emph{Computer Vision -- ECCV 2016}.\hskip 1em plus 0.5em minus 0.4em\relax Cham: Springer International Publishing, 2016, pp. 21--37.

\bibitem{multibox}
D.~Erhan, C.~Szegedy, A.~Toshev, and D.~Anguelov, ``Scalable object detection using deep neural networks,'' in \emph{Proceedings of the IEEE Conference on Computer Vision and Pattern Recognition (CVPR)}, June 2014.

\bibitem{vgg_net}
K.~Simonyan and A.~Zisserman, ``Very deep convolutional networks for large-scale image recognition,'' \emph{arXiv preprint arXiv:1409.1556}, 2014.

\bibitem{retinanet}
T.-Y. Lin, P.~Goyal, R.~Girshick, K.~He, and P.~Doll{\'a}r, ``Focal loss for dense object detection,'' in \emph{Proceedings of the IEEE international conference on computer vision}, 2017, pp. 2980--2988.

\bibitem{resnext}
S.~Xie, R.~Girshick, P.~Doll{\'a}r, Z.~Tu, and K.~He, ``Aggregated residual transformations for deep neural networks,'' in \emph{Proceedings of the IEEE conference on computer vision and pattern recognition}, 2017, pp. 1492--1500.

\bibitem{redmon2018yolov3}
J.~Redmon and A.~Farhadi, ``Yolov3: An incremental improvement,'' \emph{arXiv preprint arXiv:1804.02767}, 2018.

\bibitem{fcos}
Z.~Tian, C.~Shen, H.~Chen, and T.~He, ``Fcos: Fully convolutional one-stage object detection,'' in \emph{Proceedings of the IEEE/CVF international conference on computer vision}, 2019, pp. 9627--9636.

\bibitem{yolox}
Z.~Ge, S.~Liu, F.~Wang, Z.~Li, and J.~Sun, ``Yolox: Exceeding yolo series in 2021,'' \emph{arXiv preprint arXiv:2107.08430}, 2021.

\bibitem{yolov5}
\BIBentryALTinterwordspacing
G.~Jocher, A.~Chaurasia, A.~Stoken, J.~Borovec, NanoCode012, Y.~Kwon, K.~Michael, TaoXie, J.~Fang, imyhxy, Lorna, Z.~Yifu, C.~Wong, A.~V, D.~Montes, Z.~Wang, C.~Fati, J.~Nadar, Laughing, UnglvKitDe, V.~Sonck, tkianai, yxNONG, P.~Skalski, A.~Hogan, D.~Nair, M.~Strobel, and M.~Jain, ``{ultralytics/yolov5: v7.0 - YOLOv5 SOTA Realtime Instance Segmentation},'' Nov. 2022. [Online]. Available: \url{https://doi.org/10.5281/zenodo.7347926}
\BIBentrySTDinterwordspacing

\bibitem{detr}
N.~Carion, F.~Massa, G.~Synnaeve, N.~Usunier, A.~Kirillov, and S.~Zagoruyko, ``End-to-end object detection with transformers,'' in \emph{European conference on computer vision}.\hskip 1em plus 0.5em minus 0.4em\relax Springer, 2020, pp. 213--229.

\bibitem{vaswani2017attention}
A.~Vaswani, N.~Shazeer, N.~Parmar, J.~Uszkoreit, L.~Jones, A.~N. Gomez, {\L}.~Kaiser, and I.~Polosukhin, ``Attention is all you need,'' \emph{Advances in neural information processing systems}, vol.~30, 2017.

\bibitem{bert}
J.~Devlin, M.-W. Chang, K.~Lee, and K.~Toutanova, ``Bert: Pre-training of deep bidirectional transformers for language understanding,'' \emph{arXiv preprint arXiv:1810.04805}, 2018.

\bibitem{cordonnier2019relationship}
J.-B. Cordonnier, A.~Loukas, and M.~Jaggi, ``On the relationship between self-attention and convolutional layers,'' \emph{arXiv preprint arXiv:1911.03584}, 2019.

\bibitem{swin}
Z.~Liu, Y.~Lin, Y.~Cao, H.~Hu, Y.~Wei, Z.~Zhang, S.~Lin, and B.~Guo, ``Swin transformer: Hierarchical vision transformer using shifted windows,'' in \emph{Proceedings of the IEEE/CVF International Conference on Computer Vision}, 2021, pp. 10\,012--10\,022.

\bibitem{deformable_detr}
X.~Zhu, W.~Su, L.~Lu, B.~Li, X.~Wang, and J.~Dai, ``Deformable detr: Deformable transformers for end-to-end object detection,'' \emph{arXiv preprint arXiv:2010.04159}, 2020.

\bibitem{codetr}
Z.~Zong, G.~Song, and Y.~Liu, ``Detrs with collaborative hybrid assignments training,'' in \emph{Proceedings of the IEEE/CVF international conference on computer vision}, 2023, pp. 6748--6758.

\bibitem{rareplanes}
J.~Shermeyer, T.~Hossler, A.~Van~Etten, D.~Hogan, R.~Lewis, and D.~Kim, ``Rareplanes: Synthetic data takes flight,'' in \emph{Proceedings of the IEEE/CVF Winter Conference on Applications of Computer Vision}, 2021, pp. 207--217.

\bibitem{dota}
G.-S. Xia, X.~Bai, J.~Ding, Z.~Zhu, S.~Belongie, J.~Luo, M.~Datcu, M.~Pelillo, and L.~Zhang, ``Dota: A large-scale dataset for object detection in aerial images,'' in \emph{Proceedings of the IEEE conference on computer vision and pattern recognition}, 2018, pp. 3974--3983.

\bibitem{xview}
D.~Lam, R.~Kuzma, K.~McGee, S.~Dooley, M.~Laielli, M.~Klaric, Y.~Bulatov, and B.~McCord, ``xview: Objects in context in overhead imagery,'' \emph{arXiv preprint arXiv:1802.07856}, 2018.

\bibitem{mmdetection}
K.~Chen, J.~Wang, J.~Pang, Y.~Cao, Y.~Xiong, X.~Li, S.~Sun, W.~Feng, Z.~Liu, J.~Xu, Z.~Zhang, D.~Cheng, C.~Zhu, T.~Cheng, Q.~Zhao, B.~Li, X.~Lu, R.~Zhu, Y.~Wu, J.~Dai, J.~Wang, J.~Shi, W.~Ouyang, C.~C. Loy, and D.~Lin, ``{MMDetection}: Open mmlab detection toolbox and benchmark,'' \emph{arXiv preprint arXiv:1906.07155}, 2019.

\end{thebibliography}

\end{document}